\documentclass{article}
\usepackage{booktabs}
\usepackage{amssymb}
\usepackage{xcolor}
\usepackage{graphicx}
\usepackage{mathptmx}
\usepackage{url}
\usepackage{amssymb}

\title{TaBIIC2: Interactive Building of\\
Ontological Taxonomies using \\Weighted Self-Organizing Maps}
\author{Mathieu d'Aquin
\\ \texttt{mathieu.daquin@loria.fr} \\
\textit{MosAIK Team, LORIA, CNRS, Université de Lorraine}}
\date{May 2026}

\begin{document}

\maketitle


\begin{abstract}
Ontologies represent the conceptual knowledge of a domain. At the core of an ontology is the taxonomy of concepts and subconcepts that represent specific entities, which can be complex to build. In many cases, information is available in the form of records describing the characteristics of relevant entities, i.e., tabular data. Identifying patterns and similarities in such data can serve as a basis for identifying concepts and organizing them. However, doing so manually can be challenging, and purely automatic approaches, such as agglomerative clustering or relying on a large language model to analyze the data, can leave the user with overwhelming results and little control. In this paper, we describe a tool that enables the progressive and interactive construction of a taxonomy of concepts by identifying clusters as well as their intentional definitions. To do so, we rely on weighted self-organizing maps as a clustering method because they enable the creation of an arbitrary number of clusters that are distinct with respect to the distributions of values of specific characteristics of the clustered entities. We show that, by integrating this mechanism and others for rapidly creating concepts that group together instances from tabular data, this tool represents a middle ground between purely manual analysis and automatic methods for building ontological taxonomies.  
\end{abstract}


\thispagestyle{empty}
\pagestyle{empty}

\section{Introduction}

We illustrate the objective of the tool presented in this paper using the well-known IRIS dataset,\footnote{\url{https://archive.ics.uci.edu/dataset/53/iris}} often used as a toy example for machine learning methods. This dataset contains the dimensions (length and width) of petals and sepals for 150 iris flowers, as well as their species. We ignore the species information here. Indeed, we place ourselves in a scenario where we have obtained observations about iris flowers in the wild and would like to organize them through a taxonomy. This taxonomy could form both the starting point and the backbone of an ontology of iris flowers, defining classes of flowers based on relevant characteristics.

Doing this manually requires analyzing the collected data to understand which characteristics are most important for distinguishing subsets of flowers, which thresholds, constraints, or restrictions can define concepts from those subsets, and how different characteristics and restrictions can be used to further subdivide those concepts into subconcepts. While this can seem daunting, and possibly very subjective, for a dataset with only four characteristics and 150 flowers, it becomes insurmountable for larger tabular datasets.

Naturally, methods for automatically finding subgroups within datasets could be used to support this task. Clustering methods in particular seem relevant. However, when used automatically, such methods provide little control to the user, leaving them with arbitrary numbers of clusters that are mostly defined by similarity rather than by clear definitions based on the characteristics of the clustered objects.

In this paper, we propose a method and a tool that combine both approaches, relying on human judgment as well as an iterative and interactive use of automatic clustering. In this tool, the user can create concepts by manually defining restrictions that constrain and refine an existing concept, or by combining existing concepts (through unions, intersections, or complements). More importantly, the tool provides a way to identify relevant subconcepts of an existing concept that are defined by restrictions on key distinguishing characteristics of the instances being subdivided. To achieve this, we rely on weighted self-organizing maps (weighted SOMs, or WSOMs). SOMs are neural clustering models that place objects into units topologically organized in a (generally square) grid. Their advantage is that they already include some structure (through the grid) and do not require fixing the number of clusters in advance (only the size of the map, with unit merging to form clusters in a postprocessing step). WSOMs, in addition to learning the map for clustering, learn weights applied to the features of the clustered objects (their characteristics), so that less distinguishing features are largely ignored during clustering.

By using WSOMs interactively and iteratively, the tool can automatically identify characteristics and restrictions that can be used to divide concepts into subconcepts. Here, we describe the tool and the approach used to achieve this, as well as their application not only to the IRIS dataset\footnote{A video of the creation of the iris taxonomy/ontology with the tool is available at \url{https://youtu.be/bQVdAtW_X0s}.}, but also to a larger case study (several tens of thousands of car descriptions). We compare the resulting ontological taxonomies with methods that generate cluster taxonomies purely automatically, as well as with the use of LLMs (large language models) for the same task, showing that our tool helps produce high-quality taxonomies from tabular data with low effort while preserving user control and understanding of the output.

The code of the tool and the tests reported in this paper are available in the repository at \url{https://github.com/mdaquin/TaBIIC2}.

The remainder of the paper is organized as follows. Section~\ref{sec:relatedwork} discusses relevant related work, including automatic hierarchical clustering of tabular data. Section~\ref{sec:preliminaries} introduces notions needed to read the paper. We then describe the features of the tool, including manual taxonomy editing (Section~\ref{sec:manual}) and the WSOM-based method for discovering relevant subconcepts (Section~\ref{sec:auto}). We also show how the created taxonomy can be exported into a lightweight ontology in OWL (Section~\ref{sec:export}). Finally, Section~\ref{sec:casestudy} describes the use of the tool on a larger dataset and compares the results with those of other methods.

\section{Related work}
\label{sec:relatedwork}

Taxonomies of concepts are at the heart of ontologies, and building them therefore represents a key part of ontology construction and development. Many methodologies have focused on the (mostly manual) design and construction of ontologies (for example,~\cite{fernandez1997methontology,suarez2012neon}). In general, these methodologies rely more on expert knowledge and knowledge capture than on bottom-up approaches that let ontologies emerge from existing observational data. Because this activity is particularly time-consuming, many approaches have been proposed to (at least partially) automate the extraction of concepts and relations between concepts, especially from textual data (see, for example,~\cite{hearst1992lexical,maedche2001ontology,buitelaar2005ontology}). However, to the best of our knowledge, very few works have specifically considered ontology learning from tabular data, with most existing efforts focusing on learning from relational databases~\cite{mosca2023review}.

Despite a strong focus on manual ontology construction or ontology building from text, some attempts have used clustering or formal concept analysis to automatically extract prototypical taxonomies from tabular data. Broadly speaking, formal concept analysis (FCA)~\cite{FCA} builds concept hierarchies from formal contexts, i.e., representations of data objects with binary attributes. The hierarchy built through FCA is formally a complete lattice containing all closed itemsets. Since FCA builds concept hierarchies from structured data, it has naturally been applied multiple times to support the semi-automatic creation of taxonomies. In~\cite{contento}, for example (applied to data licenses in~\cite{licensecontento}), FCA is used to create an initial concept hierarchy from the data, which is then manually pruned and labeled to form the core of an ontology. Many other works have used FCA to build taxonomies or parts of ontologies~\cite{taxoFCA,taxoFCA2}, or to align and merge them~\cite{fcamerging,fcamerging2}. However, extracted concepts are often unlabeled, and lattice construction frequently leads to very large and complex hierarchies that must be filtered and adapted before use in ontologies.

Other approaches have relied on clustering methods~\cite{clustering}, in particular hierarchical agglomerative clustering (HAC~\cite{AC}). For example, in~\cite{taxoAC}, the authors use this approach on a dataset of design-research articles as a starting point for creating a taxonomy of design-research issues. In~\cite{taxoAC2}, a similar process is used to build a taxonomy of IT subjects. However, an obvious disadvantage of using (hierarchical) clustering for taxonomy construction is that the resulting clusters are not explicitly defined: they are formed by grouping similar objects rather than by applying intentional restrictions that characterize the set of included objects. They are therefore not straightforwardly suitable for use in ontologies.

Finally, some works have attempted to use clustering methods interactively in order to combine user knowledge, as in manual approaches, with automated support through clustering. In particular, the first version of the TaBIIC~\cite{tabiic} system enables the user to define subconcepts by successively applying the K-means algorithm to find binary partitions of concepts, and then adjusting the intentional restrictions (constraints on attribute values) of the resulting clusters. However, it is limited to initially finding only two clusters and can create only tree-like taxonomies.

\section{Preliminaries}
\label{sec:preliminaries}

We begin by defining the simple framework of concepts used to build taxonomies. The data we rely on is tabular, in the sense that it is represented as a set of \textit{rows}, each representing an entity or object and characterized by values for named \textit{columns}. In the iris example, each row represents one iris flower, with values for the PetalLength, PetalWidth, SepalLength, and SepalWidth columns. Columns can be numerical, date-valued, categorical, or string-valued.

In general, a \textit{concept} corresponds to a set of objects (rows) that share properties. In line with the vocabulary used in FCA, we call the \textit{extension} of a concept the set of objects (rows) represented by the concept, and the \textit{intension} of a concept the definition of their shared properties.

The main mechanism through which the intension of a concept can be defined is by specifying a set of restrictions:
\begin{description}
    \item[For numerical and date columns:] this corresponds to inequality conditions ($>$, $<$, $\geq$, $\leq$) on the values of a column (e.g., $PetalLength > 4.4$).
    \item[For categorical columns:] this corresponds to equality conditions ($=$) on the value of a column (e.g., $specis = virginica$).
\end{description}
In other words, creating a subconcept from a concept using one or more restrictions corresponds to creating a new concept whose extension is the subset of the parent concept's extension that fulfills the restriction(s).

In addition, concepts can be defined by combining others, i.e., concepts whose extension is the union or intersection of multiple concepts, or the complement of multiple concepts that share a parent concept.

As we will see in the next section, this framework is designed to be easily editable manually, i.e., to make it simple to create new subconcepts by defining restrictions or combining existing concepts. However, it also formally corresponds to a restricted subset of the $ALC$ description logic~\cite{baader2003description} with concrete domains, and therefore to the OWL ontology language~\cite{OWL}. For example, defining a subconcept of the Iris concept with the restriction $PetalLength < 4.4$ corresponds to defining it as
$$
Iris~\sqcap~\exists PetalLength.decimal[< 4.4]
$$
or, in OWL Manchester syntax,
$$
Iris~\textbf{and}~PetalLength~\textbf{some}~decimal[< 4.4]
$$
Similarly, the union, intersection, and complement mechanisms are directly aligned with the corresponding constructs (or, and, not) in description logic/OWL.

\section{Manually editing taxonomies from tabular data}
\label{sec:manual}

\begin{figure}[ht]
    \centering
    \includegraphics[width=0.8\textwidth]{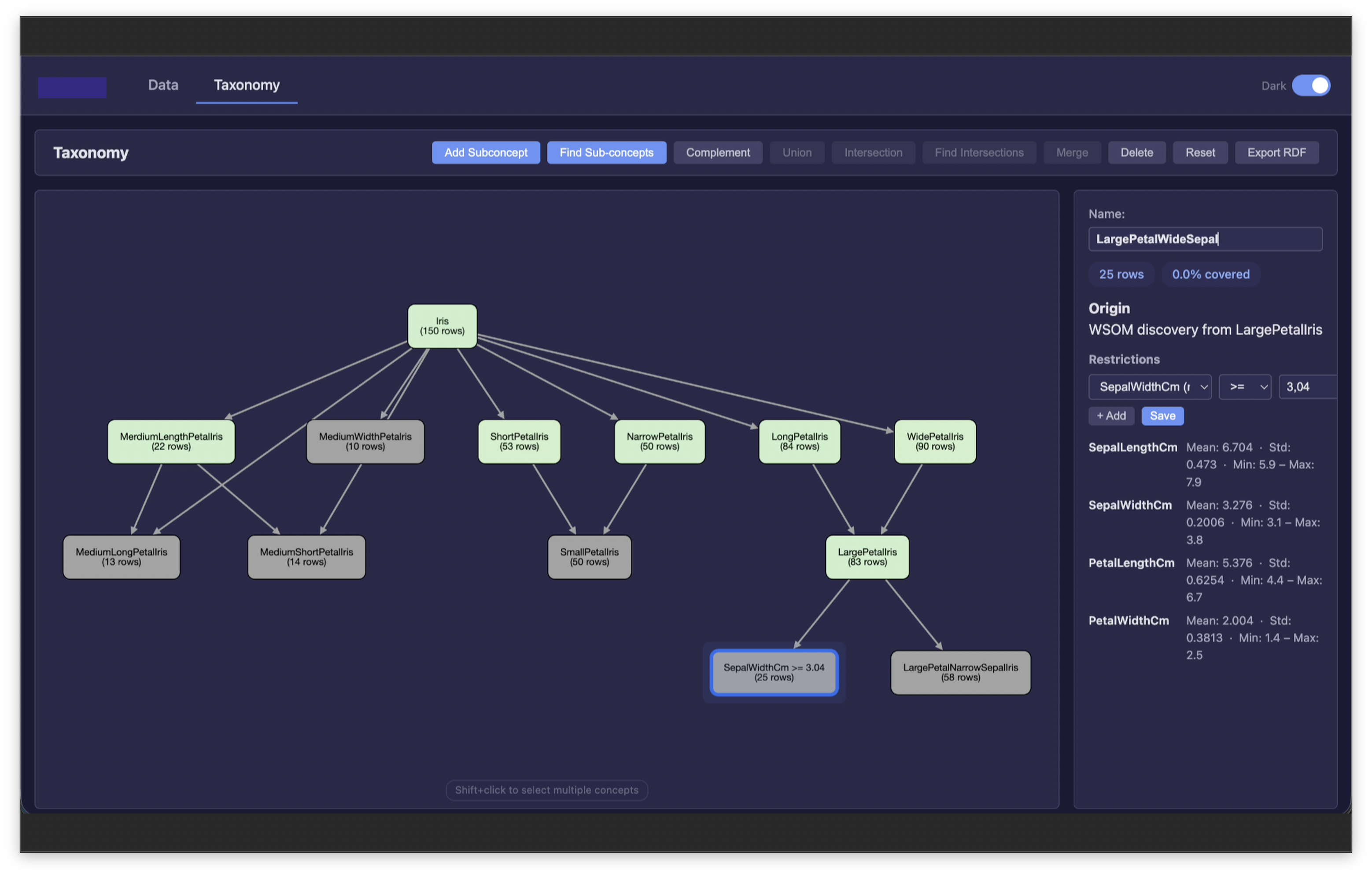}
    \caption{Screenshot of the main interface of the tool for editing taxonomies from tabular data.}
    \label{fig:screenshot}
\end{figure}

In this section, we start by provinding a summary of the features of the proposed tool to edit ontological taxonomies in the framework described in the previous section. While not the core contribution of this work, those features already suport the user in building taxonomies that are relevant to a dataset represented as tabular data, and that is already instanciated by the rows of this dataset. Figure~\ref{fig:screenshot} presents a screenshot of this editing interface with the taxonomy for the iris dataset being edited. 

The first step in the tool consists of loading and processing the data. Tabular data can be provided as a CSV or Excel (xlsx) file, where the first row includes the column names and each subsequent row includes values for those columns for one object in the dataset. The tool then identifies each column type as numerical/date-like (e.g., ``PetalLength''), categorical (e.g., ``Species''), or title-/identifier-like (e.g., ``ID''). The automatically detected type can be modified by the user if it is not correctly identified. For each column, the tool also shows basic statistical information about its values, and the user can decide to exclude it from the taxonomy-building process (for the iris example, we exclude the ID and species columns).

Once the data is loaded and processed, the user can edit the taxonomy. At the start of the process, only one concept exists: the root concept, whose extension includes the entire dataset. One or several concepts can be selected. If one concept is selected, the user can create a subconcept by defining one or more restrictions to apply to the extension of the selected concept. If multiple concepts are selected, the user can create a new concept based on the union, intersection, or complement of the selected concepts. In addition, a ``merge'' function can be used to take two or more concepts defined by restrictions and replace them with one concept whose restrictions cover those of the selected concepts. A ``find-intersections'' function can also be used to create all non-empty pairwise intersections of selected concepts.

Finally, for a selected concept, the tool enables the user to assign a human-readable label (``Iris'' for the root concept, ``LongPetalIris'' for one created with a restriction on petal length, etc.). It displays the concept's intension (restrictions or other expressions) and statistical information on column values for rows in its extension (right panel in Figure~\ref{fig:screenshot}).

\section{Automatically finding relevant subconcepts using weighted SOMs}
\label{sec:auto}

Building on the manual taxonomy framework, whose concepts are directly populated by rows from tabular data, the core contribution here is to provide support for automatically identifying relevant concepts to add to the taxonomy. The idea is that, given a selected concept in an evolving taxonomy, we should identify subsets of its extension that both represent a good partition of that extension (in clustering terms, subsets that are internally homogeneous and distinct from one another) and can be described through simple sets of restrictions. To achieve this, we rely on WSOMs. SOMs are an interesting clustering method because they can produce an arbitrary number of clusters (up to the number of units in the grid, with cluster construction achieved by post-processing and merging units). However, as with many purely similarity-based clustering techniques, the resulting clusters group \textit{generally similar} objects and do not necessarily focus on specific properties they share. WSOMs additionally learn importance weights for each feature (columns, in our case), enabling the identification of groups that are distinct with respect to a small subset of features, which in turn makes it easier to derive restrictions that characterize each cluster. In this section, we first provide a short introduction to SOMs and WSOMs, and then describe how we use WSOM training results on the extension of a concept to identify a small number of subconcepts characterized by restrictions on specific columns. In the implementation of the tool, we rely on the ksom\footnote{\url{https://pypi.org/project/ksom/}} Python library.

\subsection{Self-organizing maps}

A SOM is a neural network that usually consists of a two-dimensional grid (most often square), where each unit is connected to all units in the input layer and is associated with an activation function corresponding to a distance between the unit's weight vector and the input vector. A SOM is not trained in the usual way through backpropagation, but through competitive learning. Intuitively, for each example in the training set, a best matching unit (BMU) is first identified as the one with the lowest activation (highest similarity). Then, the weights of all units are updated using the following formula:

$$W_i' = W_i + \theta(BMU, i, \beta) \cdot \alpha \cdot (X-W_i)$$

where $W_i$ corresponds to the weights of unit $i$ in the SOM, $X$ is the input vector, $\alpha$ is the learning rate (decreasing over time), and $\theta$ is the neighborhood function applied between the BMU and unit $i$, with radius $\beta$. This neighborhood function should have value $1$ for the BMU itself and gradually decrease (depending on the decaying radius $\beta$) as unit $i$ gets farther from the BMU. 
Over multiple iterations, this creates a map in which similar data points have BMUs located close to each other, while dissimilar data points have BMUs in different locations on the map. Fundamentally, this is a clustering method, since each unit can be considered a cluster, and contiguous units that are BMUs for some data points can be merged to form larger clusters.

\subsection{Weighted self-organizing maps}

As the name indicates, WSOMs add a component to SOMs: a vector of weights associated with the input features of the map. In other words, instead of directly using a similarity function to identify the BMU, WSOM first multiplies input vectors (values of the columns, in the case of tabular data) by this weight vector, thereby assigning more or less importance to each feature.

The main idea is that WSOM training, in addition to the BMU-based process described in the previous section, also learns weights that lead to units (clusters) that are more distinctive from one another.

To achieve this, the weights are trained using a backpropagation-based process with a loss function that characterizes how much the weighted input vector is more similar to the BMU than to the other units of the map. In the library used in the tool, the following loss function is used:

$$
L(X, M) = 1 - \frac{\displaystyle\min_{M_i \in M}\bigl(\mathrm{dist}(X, M_i)\bigr)}{\frac{1}{|M|}\displaystyle\sum_{M_i \in M} \mathrm{dist}(X, M_i)}
$$

where $M$ is a tensor representing the map, $X$ is a tensor representing the inputs at a given training step, and $dist$ is the distance function used (e.g., cosine distance). Plainly put, it is based on the ratio between the minimum distance from the input to a unit in the map (i.e., the distance to the BMU) and the average distance to all units in the map.

In addition, since a goal of WSOM is to identify a small number of features that are representative of the SOM clustering, a sparsity component is also applied to the loss (here, an L2 regularization term is added to the loss function).

\subsection{Identifying relevant restrictions in WSOMs clusters}

Based on the process described above, automatically identifying relevant subconcepts of an already formed concept first consists of training a WSOM on the extension of that concept (with features corresponding to the columns in the tabular data). Several hyperparameters can be adjusted by the user, including map size and the number of training epochs, with reasonable default values suggested based on data dimensions (i.e., the size of the concept for which we are trying to find subconcepts). It is also possible to ignore some columns (e.g., if subconcepts have already been created from restrictions on those columns).

The result of this training process is a map containing a number of units (often several dozen). Each unit is represented by a vector corresponding approximately to the average column values of rows for which it is the BMU, along with a weight vector in which each weight corresponds to the importance of one column in the organization of the map.

From this, the next step is to identify a (preferably small) number of concepts defined by restrictions on a specific column over the extension of the input concept. We first select the column on which restrictions will be applied as the one with the highest weight in the WSOM. We then create restrictions on this column for every unit in the map as follows:
\begin{description}
  \item[If the column is numerical/date-valued:] Find the interval (minimum and maximum) that contains all values of this column for rows for which the unit is BMU.
  \item[If the column is categorical:] Find the most frequent value among rows for which the unit is BMU.
\end{description}
We then reassign rows from the extension of the input concept to units based on those restrictions, and remove units with empty extensions. Finally, we merge units based on overlap: if the extension of a unit $A$ contains the extension of a unit $B$, the restriction for $A$ is more general than that for $B$, so $B$ is unnecessary.

The remaining units are then integrated into the tool as subconcepts of the input concept, with their intension defined by the restrictions discovered through the process described above. The user can then accept or reject these subconcepts, or rerun the process with different parameters.

\section{Converting the taxonomy into an OWL ontology}
\label{sec:export}

The process described in the previous sections (both manual editing and automatic discovery of subconcepts with WSOMs) produces a taxonomy of concepts in which some concepts are defined by specializing their parent concept(s) through restrictions, while others are defined as unions, intersections, or complements of existing concepts. The final step is therefore to convert this taxonomy into an OWL ontology, using the correspondence between this taxonomy framework and the description-logic formalism underlying OWL, as described in Section~\ref{sec:preliminaries}. This step is important because it positions the tool's functions (interactive and semi-automatic taxonomy construction) as a starting point for a broader ontology development process. In other words, once exported to OWL, the taxonomy can be further edited in tools such as Protégé\footnote{\url{https://protege.stanford.edu/}} and refined into a proper ontology.

\section{A case study: A taxonomy of cars}
\label{sec:casestudy}

So far, we have mostly exemplified the process on the iris dataset, a small-scale tabular dataset containing only a few numerical columns. In the repository, we include the taxonomy produced\footnote{\url{https://github.com/mdaquin/TaBIIC2/blob/main/tests/TaBIIC2/iris.ttl}} using our tool, as well as taxonomies built with other methods. To further demonstrate the benefits of the tool and the method it implements, we describe here a larger and more complex case study: building a taxonomy of cars. We compare the results obtained using our tool (interactively) with taxonomies produced by three automatic approaches: HAC, FCA, and LLMs. Naturally, evaluating taxonomies is a complex task, because taxonomy quality is subjective and depends on the intended use. We therefore focus on comparing the general characteristics of the taxonomies produced by each method and discussing what these characteristics imply in practice.

\subsection{The dataset}

For this case study, we use a relatively large dataset called the ``100,000 UK used car dataset''\footnote{\url{https://www.kaggle.com/datasets/adityadesai13/used-car-dataset-ford-and-mercedes}} which, after merging and cleaning, contains descriptions of approximately 88K used cars based on scraped sales adverts. The included columns describe brand, model, production year, asking price, transmission type (automatic, manual, semi-auto), mileage, fuel type (petrol, diesel, electric, hybrid, other), tax band, fuel consumption (in miles per gallon, mpg), and engine size (in liters). For taxonomy building, we always ignore brand and model (because there are too many distinct values), as well as tax band, since it is calculated from an aggregation of several other car characteristics and is therefore redundant.

\subsection{Methods tested for building the taxonomy}

\begin{table}[ht]
\centering
\caption{Comparison of taxonomy-building approaches from tabular data.}
\label{tab:taxonomy-comparison}
\begin{tabular}{lllll}
\toprule
\textbf{Method} & \textbf{Concept interpretability} & \textbf{Automation} & \textbf{User control} & \textbf{Multi-parent support} \\
\midrule
Our tool           & explicit restrictions  & semi (WSOM)  & full              & \checkmark \\
FCA               & formal intents         & full         & ---               & \checkmark\ (lattice) \\
HAC               & approximate post-hoc   & full         & ---               & tree only \\
LLM-assisted      & natural language       & semi         & via prompting     & \checkmark \\
\bottomrule
\end{tabular}
\end{table}

Table~\ref{tab:taxonomy-comparison} provides an overview of the methods used to build taxonomies for the dataset presented in the previous section, together with their main characteristics. As described above, our tool was used while ignoring the brand, model, and tax-band columns. The ``engineSize'', ``mpg'', ``year'', ``price'', and ``mileage'' columns were treated as numerical, and all other columns as categorical (this was automatically detected by the tool and did not need to be changed). Building the taxonomy took approximately 20 minutes, with iterative back-and-forth based on subjective decisions and trial-and-error. Because this dataset is much larger than IRIS, WSOM training and map post-processing took substantially longer, especially for the first few concepts (higher in the taxonomy and therefore with larger extensions). The main criterion for deciding whether to divide a concept into subconcepts was to keep concepts balanced, i.e., to avoid leaf concepts with very large extensions (over 10,000 cars).

To create a taxonomy with HAC, we relied on the SciPy Python library. The same column processing used in our tool (standardization, encoding, and filtering) was applied. A hierarchy was built using Euclidean distance, Ward linkage, and a maximum depth of 20.\footnote{See the configuration file at \url{https://github.com/mdaquin/TaBIIC2/blob/main/tests/HAC/cars_config.json}.} However, due to the computational cost of the method (especially memory usage), it was not possible to apply it to the full dataset. We therefore used a random sample containing only 10,000 cars. Besides showing a clear limitation of HAC-based methods on larger datasets, this also affects the values reported in the next section.

We also implemented an FCA-based method, building a concept lattice with the same configuration as the other methods (in terms of selected column types, selected columns, and maximum depth). It should be noted that, although this method managed to process the full dataset, it took more than 30 minutes to finish.

Finally, we tested several commercial LLMs (ChatGPT, Gemini, Claude) on the task of creating a taxonomy from the dataset. We devised a prompt aimed at making them as effective as possible on this task, specifying which columns should be ignored and providing examples of restrictions (in OWL, since we instructed the LLMs to produce the taxonomy directly in OWL).\footnote{The prompt used can be found at \url{https://github.com/mdaquin/TaBIIC2/blob/main/tests/LLMs/cars_prompt.txt}}

\subsection{Results}

\begin{table}[h]
\centering
\caption{Taxonomy statistics for the cars dataset across different methods.}
\label{tab:cars-stats}
\resizebox{\textwidth}{!}{%
\begin{tabular}{lrrrrrrrrrrr}
\toprule
\textbf{Method}
  & \rotatebox{90}{\textbf{Concepts}}
  & \rotatebox{90}{\textbf{Instances}}
  & \rotatebox{90}{\textbf{Restrictions}}
  & \rotatebox{90}{\textbf{Levels}}
  & \rotatebox{90}{\textbf{Leaves}}
  & \rotatebox{90}{\textbf{Multi-parent}}
  & \rotatebox{90}{\textbf{Avg branching}}
  & \rotatebox{90}{\textbf{Std branching}}
  & \rotatebox{90}{\textbf{Avg instances}}
  & \rotatebox{90}{\textbf{Avg leaf instances}} \\
\midrule
HAC                & 19{,}858 &  10{,}015 &   0  & 21 &  9{,}929 &     0 & 2.000 & 0.010 &      0.5 &      1.0 \\
FCA                &  9{,}677 &  88{,}076 & 47{,}461 & 8 &  949 & 9{,}652 & 5.194 & 3.230 &      9.1 &     92.8 \\
ChatGPT 5.2        &    66 &  88{,}068 &  65     &  4 &     45 &     0 & 3.095 & 0.436 &  1{,}334.2 &  1{,}956.9 \\
Claude Haiku 4.5   &    20 &  88{,}068 &  19     &  4 &     13 &     0 & 2.714 & 0.488 &  4{,}403.0 &  6{,}568.8 \\
Claude Opus 4.6    &     7 &  88{,}068 &   6     &  3 &      4 &     0 & 2.000 & 0.000 & 12{,}580.0 & 22{,}015.0 \\
Claude Sonnet 4.6  &    12 &  88{,}068 &  11     &  3 &      8 &     0 & 2.750 & 0.500 &  7{,}338.3 & 11{,}007.5 \\
Gemini 3 Fast      &     7 &   1{,}008 &   6     &  3 &      4 &     0 & 2.000 & 0.000 &    142.9 &    250.0 \\
Gemini 3 Reasoning &    16 &  88{,}068 &  15     &  3 &     10 &     0 & 2.500 & 1.225 &  5{,}503.8 &  8{,}806.0 \\
Our tool           &    38 &  88{,}243 &  28     & 6 &    24 &     9 & 3.286 & 0.994 & 12{,}458.5 &  4{,}036.9 \\
\bottomrule
\end{tabular}
}
\end{table}

Table~\ref{tab:cars-stats} provides an overview, through a few statistical indicators, of the taxonomies produced by the different methods described in the previous section. All these results are available as OWL ontologies in the code repository and can therefore be inspected directly.

Starting with HAC, this method produced a very large taxonomy with close to 20K concepts, reaching the maximum depth we had set, even though it only operated on a filtered dataset (10K rows in input tabular data). Many of these concepts are leaves, each covering a single row. Also, due to the nature of the method, each intermediary concept always has exactly two subconcepts and each concept has only one parent (i.e., the taxonomy is a tree). Finally, because of how this method works, concepts are neither named in a meaningful way nor defined by restrictions. In other words, once built, this taxonomy would require substantial reengineering effort to be exploitable.

The conclusion is similar for the taxonomy produced using FCA. It is also quite large (10K concepts) and deep (8 levels). Contrary to HAC, however, intermediary concepts can have many subconcepts, and concepts can generally have many parent concepts. In other words, FCA builds a hierarchy (a lattice) with a rather complex structure. In this case, concepts are also not named, but they are defined by restrictions derived from the attributes of the itemsets from which the concepts are built. Inspecting these restrictions\footnote{See \url{https://github.com/mdaquin/TaBIIC2/blob/main/tests/FCA/cars.ttl}.}, one can observe that they are often quite complex (multiple restrictions mixing multiple properties), which makes the resulting concepts difficult to interpret.

In contrast, on average, the LLMs tested produced relatively simple taxonomies. With the exception of ChatGPT~5.2, all taxonomies created in this way contain only a few concepts, each defined by one restriction. Concepts always have one parent (so the structure is a tree rather than a general hierarchy) and only a few subconcepts, leading to a fairly balanced and simple structure. Concept names are generally consistent with the restrictions defining them, which are themselves usually simple. It can nevertheless be observed that all LLMs created restrictions with thresholds that do not really fit the provided data, and instead tend toward intuitive cut-off points such as ``10,000''.

The taxonomy created with our tool is one of several that could have been produced, and it naturally reflects user knowledge and choices. It remains simple, while still being more sophisticated than those created by most LLMs. The structure includes multiple parents and arbitrary numbers of subconcepts but, thanks to the focus on user control, remains balanced and interpretable. The created restrictions are also relatively simple and are designed to fit the data directly, i.e., to create concept subdivisions that emerge from the data rather than from arbitrary cut-off points. As a result, this initial taxonomy, created with user input but in a time comparable to that required by the other methods (especially considering the dataset scale), represents a good starting point for refinement into an ontology of used cars.

\section{Conclusion}

In this paper, we presented a tool that combines advanced automatic clustering with user control and manual input to create taxonomies from tabular datasets. Through a relatively large-scale case study (88K data points), we showed that this approach can produce sophisticated yet balanced and interpretable taxonomies, which can then be further refined using ontology development tools. In particular, by comparison with more automated methods, we showed that, for this task, combining well-integrated automation with human judgment can lead to more directly exploitable results while reducing user cognitive effort. This is supported, at least in part, by the use of WSOMs (weighted self-organizing maps), a clustering method designed to produce clusters that are clearly distinct with respect to a reduced subset of input characteristics.

As with any complex tool, there are naturally several aspects that could be improved. One of them is WSOM parameterization, which can remain obscure for users. In the current implementation, reasonable defaults and values automatically set through heuristics based on input dimensions are provided. However, if those suggested values do not lead to reasonable results, understanding the impact of parameters such as map size or number of training epochs may require time. One possible direction would be to propose multiple solutions with different parameterizations and let the user choose, but this may be unrealistic given the computational and cognitive costs involved. Computational cost is itself another area for improvement. At present, the tool requires the dataset to be loaded in memory at all times, which can be a constraint for large datasets. In addition, both WSOM training and post-processing are computationally intensive, which currently restricts practical use to scales of at most a few tens of thousands of rows (as in our case study). Addressing these limitations is therefore part of our future work.

\bibliographystyle{plain}
\bibliography{tabiic2}

\end{document}